\documentclass{article}

\usepackage{arxiv}

\usepackage[utf8]{inputenc} 
\usepackage[T1]{fontenc}    
\usepackage{hyperref}       
\usepackage{url}            
\usepackage{booktabs}       
\usepackage{amsfonts}       
\usepackage{nicefrac}       
\usepackage{microtype}      
\usepackage{lipsum}		
\usepackage{graphicx}
\usepackage{natbib}
\usepackage{doi}

\usepackage{multirow}

\title{SM70: A Large Language Model for Medical Devices
\thanks{We would like to acknowledge Spass Inc., South Korea for funding this research.}
}

\date{} 					

\author{
{Anubhav Bhatti, Surajsinh Parmar, San Lee} \\
{\textit{AI Engineering Team, SpassMed Inc.}, Toronto, Ontario, Canada} \\
(anubhav.bhatti, suraj.parmar, sanlee)@spassmed.ca
}

\renewcommand{\headeright}{Technical Report}
\renewcommand{\undertitle}{Technical Report}
\renewcommand{\shorttitle}{SM70: LLM for Medical Devices}

\hypersetup{
pdftitle={SM70: A Large Language Model For Medical Devices},
pdfsubject={q-bio.NC, q-bio.QM},
pdfauthor={SPASSMED INC.},
pdfkeywords={Large Language Model, Llama2, LoRA, USMLE.},
}

\begin{document}
\maketitle

\begin{abstract}
We are introducing SM70, a 70 billion-parameter Large Language Model that is specifically designed for SpassMed's medical devices under the brand name 'JEE1' (pronounced as G1 and means 'Life'). This large language model provides more accurate and safe responses to medical-domain questions. To fine-tune SM70, we used around 800K data entries from the publicly available dataset MedAlpaca. The Llama2 70B open-sourced model served as the foundation for SM70, and we employed the QLoRA technique for fine-tuning. The evaluation is conducted across three benchmark datasets - MEDQA - USMLE, PUBMEDQA, and USMLE - each representing a unique aspect of medical knowledge and reasoning. The performance of SM70 is contrasted with other notable LLMs, including Llama2 70B, Clinical Camel 70 (CC70), GPT 3.5, GPT 4, and Med-Palm, to provide a comparative understanding of its capabilities within the medical domain. Our results indicate that SM70 outperforms several established models in these datasets, showcasing its proficiency in handling a range of medical queries, from fact-based questions derived from PubMed abstracts to complex clinical decision-making scenarios. The robust performance of SM70, particularly in the USMLE and PUBMEDQA datasets, suggests its potential as an effective tool in clinical decision support and medical information retrieval. Despite its promising results, the paper also acknowledges the areas where SM70 lags behind the most advanced model, GPT 4, thereby highlighting the need for further development, especially in tasks demanding extensive medical knowledge and intricate reasoning.
\end{abstract}


\section{Introduction}
A key advantage of these Large Language Models (LLMs) is their capacity to carry out tasks based on natural language instructions, removing the need for users to be skilled in programming. This aspect is particularly beneficial for medical professionals, enabling them to navigate and control the model across various medical processes easily. Possible uses include helping medical staff with note-taking, drafting discharge letters, extracting information from lengthy documents, summarizing content, and transforming unstructured text into organized formats, assuming the model has been adequately trained on a vast array of medical documents. These LLMs can be advantageous in medical education, acting as a study aid for students by offering quizzes or explaining complex topics, provided the model's coherence and accuracy are reliable. However, the most advanced LLMs are not freely available; they are accessible only through APIs that require data to be sent to the parent company for processing. Given the sensitive nature of medical data and the need for stringent privacy measures, models that lack transparency and have unclear data management practices are unsuitable for medical use. To address these issues and prevent unauthorized data transfers, open-source models that can be implemented on-site, thereby alleviating privacy concerns, are vital. In response to this need, we introduce a SM70, a large language model fine-tuned for biomedical tasks dedicated to support SpassMed's medical devices under the brand name 'JEE1' (pronounced as 'G1' and means 'Life'). To assess the effectiveness of the model, we evaluate its performance on (1) the United States Medical Licensing Examination (USMLE), a standardized assessment undertaken by medical students in the United States as part of their qualification process to become physicians, (2) MEDQA – USMLE style questions answers on medical domain, and (3) PubMEDQA datasets.

\section{Methodology}

\subsection{Method}
For developing SM70 large language model, we used Llama 2 70B as the foundation model \cite{touvron2023llama}. 
The primary dataset utilized for this purpose was the extensive Medical Meadow dataset, comprising approximately 800,000 data entries. This dataset encompasses a wide array of medical texts and has been previously employed by the MedAlpaca team \cite{han2023medalpaca,savery2020question,wang-etal-2020-cord,hendrycks2021ethics,jin2020disease,yu-etal-2019-detecting}. The dataset's diversity and depth provide a robust basis for training the model across various medical knowledge domains and applications.

To fine-tune the SM70 model effectively, we employed the Parameter-Efficient Fine-Tuning (PEFT) technique, specifically the QLoRA approach \cite{hu2021lora,dettmers2023qlora}. This approach enables the model to be fine-tuned in 4-bit precision, significantly reducing the computational load without compromising the model’s performance. The integration of Low Rank Adapters (LoRA) within this framework further enhances the model's learning capability, allowing it to adapt more efficiently to the specific requirements of medical question-answering tasks.

The fine-tuning process was meticulously designed to optimize the model’s performance. The hyperparameters chosen for PEFT and model training were critical in achieving the desired balance between efficiency and effectiveness. These parameters were carefully selected based on preliminary trials and validations to ensure they were well-suited for the task at hand. The details of these hyperparameters, including learning rate, batch size, number of training epochs, and other relevant settings, are systematically presented in Table \ref{tab:params}. This table serves as a comprehensive guide to the technical specifications and configurations employed in the training and development of the SM70 model.


{
\renewcommand{\arraystretch}{1.4}

\begin{table}[htbp]
\small
\caption{Training Parameters for fine-tuning SM70}
\begin{center}
\begin{tabular}{|c|c|}
\hline
\textbf{Parameters} & \textbf{SM70}\\ \hline 
Sequence Length & 1024 \\ \hline
LoRA - 'r' & 64 \\ \hline
LoRA - 'Alpha' & 16 \\ \hline
LoRA - 'Dropout' & 0.1 \\ \hline
LoRA - 'Target Modules' & All Linear Layers \\ \hline
Gradient Accumulation Steps & 1 \\ \hline
Mini-batch Size & 32 \\ \hline
Number of Epochs & 5 \\ \hline
Optimizer & ADAMW\\ \hline
Learning Rate Scheduler & Constant \\ \hline
Learning Rate & 2e-4 \\ \hline
\end{tabular}
\label{tab:params}
\end{center}

\end{table}}

\subsection{Dataset Description} 
For this work, we used the publicly available Medical Meadow dataset for fine-tuning base Llama2 70B model. The Medical Meadow dataset \cite{han2023medalpaca} is a collection of the following datasets:

\begin{itemize}
    \item MEDIQA: MEDIQA is a dataset of manually generated, question-driven summaries of multi and single document answers to consumer health questions \cite{savery2020question}.
    \item MMMLU: The dataset Measuring Massive Multitask Language Understanding (MMMLU) \cite{hendrycks2020measuring} presents a comprehensive study on language understanding across a wide range of tasks. The research focuses on evaluating language models' performance in a multitask setting, demonstrating their capabilities and limitations in understanding complex language scenarios.
    \item Anki Flashcards: The Anki Medical Curriculum flashcards, crafted and regularly updated by medical students, encompass the full scope of the curriculum. They include topics like anatomy, physiology, pathology, pharmacology, among others, and often contain concise summaries and memory aids to facilitate the learning and memorization of essential medical principles.
    \item Wikidoc Patient Information: A medical question-answer pairs extracted from WikiDoc, a collaborative platform for medical professionals to share and contribute to up-to-date medical knowledge. Patient Information is structured differently, in that each section subheading is already a question, making rephrasing them obsolete.
    \item Wikidoc Living Textbook: The "Living Textbook" from WikiDoc contains chapters for various medical specialties. The dataset was extracted  using GTP-3.5-Turbo to rephrase the paragraph heading to a question and used the paragraph as answer.
    \item Pubmed Causal: The PubMed Causal dataset \cite{yu-etal-2019-detecting}, featured in the paper "Detecting Causal Language Use in Science Findings," is designed to facilitate the study of causal language in scientific texts, particularly in the realm of health and medicine. It serves as a resource for training and evaluating natural language processing models on their ability to detect and understand causal relationships in scientific literature.
    \item MEDQA: The MEDQA dataset \cite{jin2020disease} is a resource specifically designed for evaluating natural language processing models in the medical domain, focusing on medical question answering. The dataset's collection involved three main sources: the United States, Mainland China, and the Taiwan District. The questions and answers (QAs) from these regions are organized in separate folders, with the data presented in JSONL file format.
    \item Pubmed Health Advice: The PubMed Health Advice dataset \cite{yu-etal-2019-detecting}, utilized in the paper "Detecting Causal Language Use in Science Findings," is specifically designed for analyzing and understanding the use of causal language in scientific texts.
    \item CORD-19: CORD-19 \cite{wang-etal-2020-cord} is a resource of over 1,000,000 scholarly articles, including over 400,000 with full text, about COVID-19, SARS-CoV-2, and related coronaviruses. This freely available dataset is provided to the global research community to apply recent advances in natural language processing and other AI techniques to generate new insights in support of the ongoing fight against this infectious disease \cite{wang-etal-2020-cord}.
\end{itemize}

\subsection{Data Preprocessing}
In the development of the SM70 Large Language Model, a critical step was the amalgamation and standardization of the various datasets mentioned in the previous sections. This process involved the integration of multiple datasets to construct a unified question-and-answer (QA) dataset, suitable for training the model. The datasets incorporated a range of contexts, questions, and answers, which required systematic reformatting to ensure consistency and compatibility with our model training approach. The primary task in this preprocessing phase was to reformat the data to adhere to a specific prompt template. This template was designed to standardize the structure of the data, making it more conducive to effective model training. For datasets containing distinct sections of context, question, and answer, we combined the context and question elements to form a cohesive prompt. This amalgamation was done in a manner that preserved the flow and relevance of the information, ensuring that the resulting prompt was coherent and informative.

In addition to combining contexts and questions, we reformatted the answers to align with multiple answer type options. This step was crucial in accommodating the varied nature of responses found in the datasets, ranging from single-word answers to more elaborate explanations. The standardized format for answers ensured that the model could adequately process and learn from a wide spectrum of response types. The specific prompt template utilized for fine-tuning the SM70 model was structured as follows:
"\textit{Question: {Context} {Instruction/Question}. Answer: {Output}}". In this template, the "\textit{Question}" segment included the combined context and instruction or question from the dataset. The "\textit{Answer}" segment encapsulated the corresponding response or output. This structured approach allowed for a more streamlined and effective training process, as the model could easily parse and understand the format of the data being fed into it.


\section{Results and Discussion}
In the present analysis, we examine the performance of various Large Language Models (LLMs), including our proposed model "SM70," across three established medical question-answer datasets, namely MEDQA - USMLE, PUBMEDQA, and USMLE. These datasets are integral in assessing the proficiency of LLMs in terms of medical knowledge assimilation and question-answering abilities.

MEDQA - USMLE: In this dataset, SM70 registers a score of 60.80. This performance surpasses that of Llama2 70B (57.3), CC70 (60.7) \cite{toma2023clinical}, and GPT 3.5 (53.6), while trailing behind GPT 4 (81.4) and Med-Palm (79.7). The MEDQA - USMLE dataset, reflecting the nature of questions in the United States Medical Licensing Examination, serves as a benchmark for evaluating a model's grasp of medical knowledge and reasoning. The score achieved by SM70 in this context is indicative of its robust understanding and application of medical knowledge.

PUBMEDQA: Here, SM70 achieves a score of 77.3, slightly outperforming Llama2 70B (76.0), GPT 4 (74.4) and GPT 3.5 (60.2), and closely rivaling CC70 (77.9). However, it is marginally outperformed by Med-Palm (79.2). PUBMEDQA, known for its focus on fact-based questions derived from PubMed abstracts, is a crucial dataset for assessing the capability of a model in handling fact-oriented medical queries. The performance of SM70 in this dataset underscores its effectiveness in processing and responding to fact-based medical inquiries.

USMLE: In the USMLE dataset, SM70 records a score of 68.5, significantly improving upon the scores of Llama2 70B (64.1), CC70 (64.3), and GPT 3.5 (58.5), while not reaching the level of GPT 4 (86.6). This dataset tests a model's proficiency in clinical knowledge and decision-making skills, and the results suggest that SM70 has a commendable understanding of clinical concepts and the ability to support informed medical decisions.

{
\renewcommand{\arraystretch}{1.4}

\begin{table*}[tbp]
\small
\caption{Evaluation of Large Language Models on three publicly available medical question-answer datasets.}
\begin{center}
\begin{tabular}{|c|c|c|c|c|c|c|}
\hline
\multirow{2}{*}{\textbf{Evaluation Dataset}} & \multicolumn{6}{c|}{\textbf{Large Language Models}}\\ \cline{2-7} 

 & Llama2 70B & SM70 (Ours) & CC70 \cite{toma2023clinical} & GPT 3.5 & GPT 4 & Med-Palm  \\ \hline
MEDQA - USMLE & 57.3 & 60.8 & 60.7 & 53.6 & 81.4 & 79.7 \\ \hline
PUBMEDQA & 76.0 & 77.3 & 77.9 & 60.2 & 74.4 & 79.2 \\ \hline
USMLE & 64.1 & 68.5 & 64.3 & 58.5 & 86.6 & - \\ \hline 

\end{tabular}
\label{tab:results}
\end{center}
\end{table*}}

\section{Conclusion and Future Work}
In summary, our model, SM70, exhibits a strong and consistent performance across all three medical question-answer datasets. This demonstrates its potent capabilities in comprehending and applying medical knowledge. Particularly notable are its results in the USMLE and PUBMEDQA datasets, suggesting its suitability for tasks such as clinical decision support and medical information retrieval. Despite outperforming several models, SM70 does not yet match the performance of the leading model, GPT 4, highlighting potential areas for further refinement, particularly in aspects requiring deep medical knowledge and sophisticated reasoning.

\bibliographystyle{unsrtnat}
\bibliography{reference}  






\end{document}